\documentclass[conference]{IEEEtran}
\usepackage{lipsum} 
\usepackage{array}
\usepackage{svg}
\usepackage{amsmath}
\usepackage{amsmath,amssymb}

\usepackage{algorithm}

\usepackage{xcolor}

\usepackage{cite}
\usepackage[super]{nth}

\usepackage{algorithm}
\usepackage{algpseudocode}
\usepackage{gensymb}

\makeatletter
\newcommand\fs@norules{\def\@fs@cfont{\bfseries}\let\@fs@capt\floatc@ruled
  \def\@fs@pre{}%
  \def\@fs@post{}%
  \def\@fs@mid{\kern3pt}%
  \let\@fs@iftopcapt\iftrue}
\makeatother
\floatstyle{norules}
\restylefloat{algorithm}

\ifCLASSOPTIONcompsoc
  \usepackage[caption=false,font=normalsize,labelfont=sf,textfont=sf]{subfig}
\else
  \usepackage[caption=false,font=footnotesize]{subfig}
\fi

\usepackage{url}

\begin{document}


\title{Adversarially Diversified Rehearsal Memory (ADRM): Mitigating Memory Overfitting Challenge in Continual Learning}
\author{\IEEEauthorblockN{Hikmat Khan}
\IEEEauthorblockA{Rowan University\\
Glassboro, New Jersey, USA\\
khanhi83@rowan.edu}
\and
\IEEEauthorblockN{Ghulam Rasool}
\IEEEauthorblockA{Moffitt Cancer Center\\
Tampa, Florida, USA\\
ghulam.rasool@moffitt.org}
\and
\IEEEauthorblockN{Nidhal Carla Bouaynaya}
\IEEEauthorblockA{Rowan University\\
Glassboro, New Jersey, USA\\
bouaynaya@rowan.edu}
}
\maketitle

\begin{abstract}
Continual learning (CL) focuses on learning non-stationary data distribution without forgetting previous knowledge. Rehearsal-based approaches are commonly used to combat catastrophic forgetting. However, these approaches suffer from a problem called ``rehearsal memory overfitting'', where the model becomes too specialized on limited memory samples and loses its ability to generalize effectively. As a result, the effectiveness of the rehearsal memory progressively decays, ultimately resulting in catastrophically forgetting the learned tasks.

To address the memory overfitting challenge, we introduce the \emph{Adversarially Diversified Rehearsal Memory}, or ADRM, a novel method designed to enrich memory sample diversity and bolster resistance against natural and adversarial noise disruptions. ADRM employs the Fast Gradient Sign Method (FGSM) to introduce adversarially modified memory samples, achieving two primary objectives: enhancing memory diversity and fostering a robust response to continual feature drifts in memory samples.

We conducted extensive experiments on the CIFAR10 dataset and found that ADRM outperforms several existing CL approaches and performs comparable to state-of-the-art methods. Additionally, we demonstrated ADRM's ability to enhance CL model robustness under natural and adversarial conditions by using CIFAR10-C and adversarially perturbed CIFAR10 datasets. 

Our contributions are as follows: Firstly, ADRM addresses overfitting in rehearsal memory by employing FGSM to diversify and increase the complexity of the memory buffer. Secondly, we demonstrate that ADRM mitigates memory overfitting and significantly improves the robustness of CL models, which is crucial for safety-critical applications. Finally, our detailed analysis of features and visualization demonstrates that ADRM mitigates feature drifts in CL memory samples, significantly reducing catastrophic forgetting and resulting in a more resilient CL model. Additionally, our in-depth t-SNE visualizations of feature distribution and the quantification of the feature similarity further enrich our understanding of feature representation in existing CL approaches. Our code is publically available at https://github.com/hikmatkhan/ADRM.

\end{abstract}

\section{Introduction}
\begin{figure}
\centering
\includegraphics[width=0.49\textwidth]{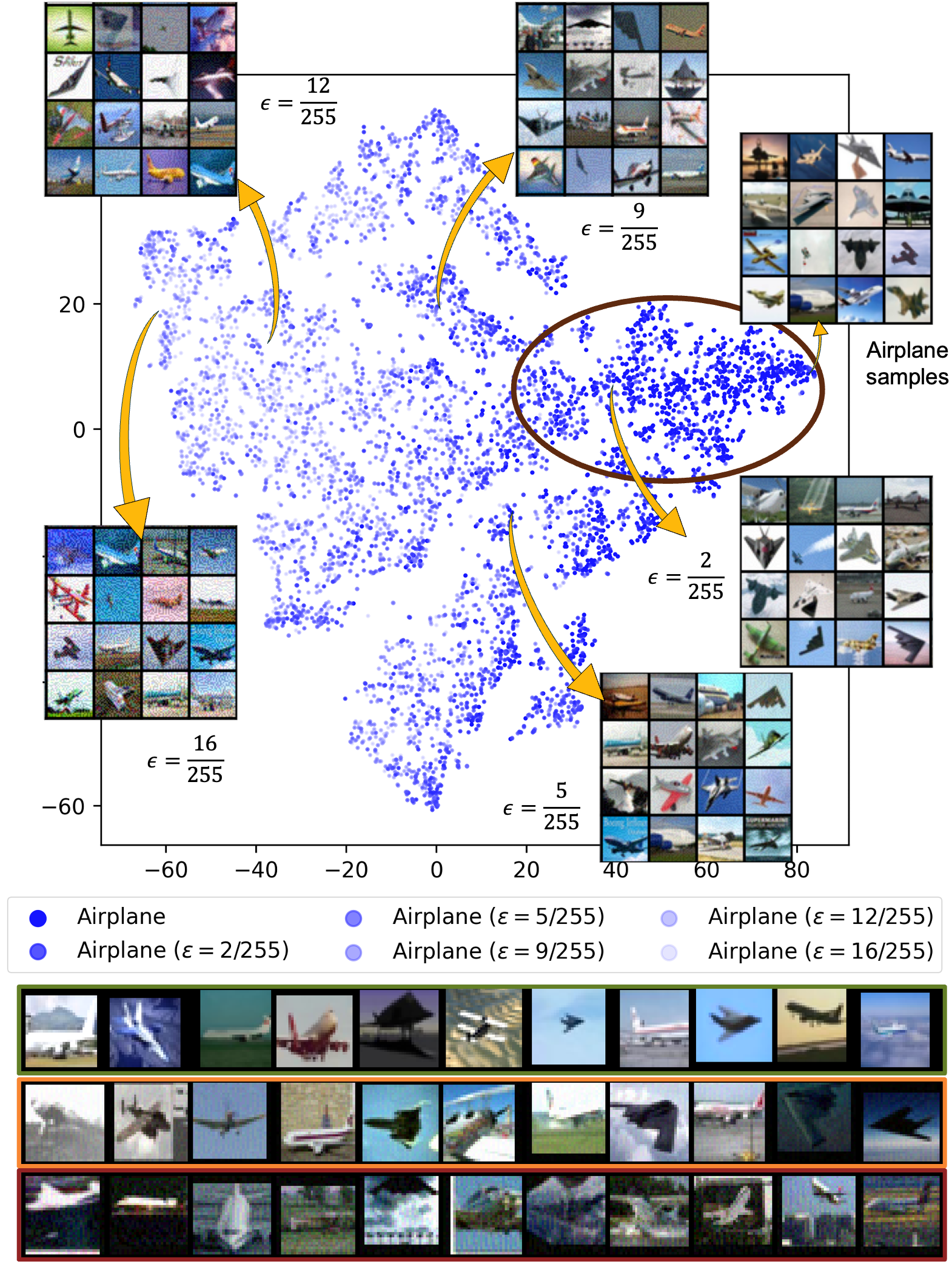}
\caption{Idea illustration:  t-SNE visualization showing the dispersion of 'airplane' class features under increasing adversarial perturbations (scaled as $\frac{\epsilon}{255}$). Dark blue dots represent the original airplane class samples, while progressively lighter shades of blue indicate the five levels of adversarially diversified samples. The visualization demonstrates how increasing the strength of the perturbation results in a divergence from the original class cluster. These adversarially diversified samples retain inherent airplane characteristics and can considered as potential rehearsal samples to enhance the diversity and complexity of the memory to prevent rehearsal memory overfitting. \nth{1} row, the samples enclosed in a green rectangle represent the original, undiversified baseline memory. \nth{2} row, the samples enclosed in an orange rectangle are adversarially diversified airplanes correctly classified by the CL model despite diversification. \nth{3} row, the samples enclosed in a red rectangle are adversarially diversified airplanes incorrectly classified, highlighting potential areas for enhancing model robustness. The diversity in rehearsal samples is crucial to prevent memory overfitting and increase the complexity of memory samples, consequently improving the robustness of the CL model (best viewed in color).}
\label{fig:main}
\end{figure}

Deep learning has exceeded human-level performance in various domains, including computer vision, natural language processing/understanding, aviation safety enhancement, reinforcement learning, and medical image analysis \cite{wang2023comprehensive,khan2020cascading, Hikmat_75, Hikmat_76, Hikmat_77, khan2020comprehensive, shah20202d}. Continual Learning (CL), also called incremental, lifelong, or sequential learning, equips deep learning models with the ability to accumulate and expand knowledge over time, similar to humans \cite{wang2023comprehensive}. Despite improvements in CL methodologies,  current approaches still suffer from a phenomenon known as \emph{catastrophic forgetting} or \emph{catastrophic interference} \cite{de2021continual, mccloskey1989catastrophic}, where models forget previously acquired knowledge after acquiring new knowledge or skills. Several approaches have been proposed to mitigate catastrophic forgetting in CL, broadly categorized into four types: 1). Regularization-based approaches introduce the penalties on alterations to crucial parameters of the model to prevent previously acquired knowledge \cite{wang2023comprehensive, de2021continual}. 2). Dynamic architecture-based approaches employ sub-networks or expandable architectures designed to adapt to new tasks while minimally impacting previously acquired knowledge \cite{wang2023comprehensive, de2021continual}. 3). Rehearsal-based approaches involve storing a subset of previous data in rehearsal memory and periodically revisiting examples from past tasks to assist the model in retaining older knowledge \cite{khan2022adversarially, rolnick2019experience, khan2024brain, khan2024}. 4). Hybrid approaches combine and aim to leverage the strengths of different learning strategies to propose the CL model that suffers less from catastrophic forgetting \cite{khan2024, wang2023comprehensive, khan2023importance, khan2022susceptibility, khan2024brain}. Rehearsal-based approaches are becoming more popular in CL approaches. These approaches use a rehearsal memory to store and rehearse the limited data samples from previously learned tasks.  However, these approaches have some limitations. The efficacy of the stored samples decreases or diminishes over time due to the limited budget of memory and lack of diversity in the stored samples \cite{de2021continual, wang2023comprehensive, khan2022susceptibility}, which potentially leads to overfitting of the stored samples \cite{khan2024, de2021continual, wang2023comprehensive, khan2024brain}. To tackle this challenge, the authors in \cite{wang2022make} proposed a continuous and reversible memory transformation function that ensures memory data is resistant to overfitting and injected the uncertainty in the transformation function, thereby enhancing the diversity of the transformed memory buffer and improving generalization. The authors in \cite{chen2023consistent} proposed an approach, called ConPL, which incorporated the previous class's prototype-based memory enhancement, resulting in less memory sample overfit and improved performance. The authors of \cite{shi2023prototype} dynamically reshape old class feature distributions by incorporating previous class prototypes with arriving new class features, preserving prior task decision boundaries. The authors in \cite{wang2022improving} dynamically evolve the distribution of memory data to improve robustness and prevent memory overfitting. Similarly, the authors in \cite{ye2022learning} proposed an evolved mixture model that adapts to data distribution shifts of the prior tasks. The authors in \cite{jin2020gradient} adopted a distinct strategy that utilized gradient-based editing of memory samples. This approach aimed to reduce memory overfitting and forgetting while incorporating updated information. The authors in \cite{bang2021rainbow} proposed a strategy called Bang's Rainbow Memory (RM) to diversify memory samples and avoid memory overfitting. This was achieved by leveraging per-sample classification uncertainty and data augmentation. Similarly, the author in \cite{tang2022learning} introduced diversifying exemplars using a learnable feature generator and semantic contrastive learning.

In this paper, we adopt a relatively straightforward approach, termed Adversarially Diversified Rehearsal Memory (ADRM), which aims to diversify and enhance the complexity of the rehearsal memory sample, which is achieved by employing the one-step Fast Gradient Sign Method (FGSM) attack on the memory samples. We then rehearse a mixture of both successful (misclassified) and unsuccessful (correctly classified) adversarially perturbed/diversified memory samples with the current task mini-batch (see Figure \ref{fig:main}). Even after adversarial perturbation, correctly classified sample options are crucial in diversifying the memory samples. Meanwhile, the misclassified samples reveal features more closely associated with the feature distributions of other memory classes or current task classes. This provides a unique opportunity for the CL model to understand the decision boundaries among different tasks. Figure \ref{fig:main} illustrates the adversarially diversified memory samples for the airplane class, showing that as the intensity of the FGSM attack increases, the resultant samples progressively deviate from the original airplane feature distribution (or core cluster). The ADRM uses FGSM to generate such diverse memory samples to effectively tackle the feature distributional drift of memory classes, as depicted in Figure \ref{fig:t_sim_viz}. We conducted extensive experiments on the CIFAR10 dataset and demonstrated that ADRM's performance outperforms several well-established CL approaches and is comparable to the state-of-the-art CL approaches. Additionally, we conducted comprehensive evaluations of ADRM and state-of-the-art baseline methods to demonstrate its stable feature distribution and robustness in natural and adversarial environments, using the CIFAR10-C \cite{hendrycks2019benchmarking} and an adversarially perturbed CIFAR10 dataset \cite{khan2022adversarially}. Lastly, we analyzed feature distribution and similarity in existing CL approaches in detail using t-SNE visualizations and Central Kernel Alignment (CKA), respectively. This analysis helped us to gain a better understanding of the learned feature representation in CL approaches. Our findings revealed that the learned feature representations in existing CL approaches were more similar to ADRM on the standard CIFAR10 dataset. However, when tested under natural and adversarial conditions, all existing CL approaches demonstrated degraded performance. Interestingly, CL approaches that suffered less forgetting showed higher similarity in learned feature representation to our ADRM, indicating the existence of continually robust features.

Our contributions are fourfold: 

\begin{itemize}
    \item To address overfitting in rehearsal memory, we proposed ADRM, utilizing the FGSM attack to diversify and enhance the complexity of the memory samples, thereby preventing memory overfit. This approach is based on the premise that robust representations aid generalization to out-of-distribution data \cite{engstrom2019adversarial}.
    
    \item We conducted extensive evaluations of ADRM and state-of-the-art methods under natural and adversarial noise conditions, demonstrating that ADRM prevents memory overfitting and enhances the robustness of the CL model, which is crucial for safety-critical scenarios.
    
    \item Through t-SNE visualization \cite{van2008visualizing} of learned feature distributions, we observed that adversarial diversification of the memory helps mitigate feature distribution drifts in memory samples compared to existing CL approaches, resulting in reduced catastrophic forgetting and a more robust CL model. 
    
    \item Our findings indicate that existing CL approaches demonstrate higher feature similarity scores (using central kernel alignment (CKA) \cite{kornblith2019similarity}) in learned features representation to the learned features of ADRM on standard CIFAR10. In contrast, they showed lesser similarity scores to the learned features of ADRM under natural and adversarial noisy conditions. These observations highlight that ADRM learned continually-adversarially robust features to mitigate catastrophic forgetting, unlike existing CL approaches, which suffered from catastrophic forgetting in natural noise and adversarial conditions.

\end{itemize}
This section covers the motivation behind the adversarial diversification of memory used in ADRM, which is explained in Section \ref{sec:motivation}. The ADRM is based on the conventional rehearsal, a widely used strategy in Continual Learning, which is elaborated in Section \ref{sec:conventional_memory_replay}. We then present our approach in detail, which involves two steps: creating the adversarially diversified memory through FGSM perturbations, discussed in Section \ref{sec:fgsm_diversification}, and rehearsing the diversified memory samples using a strategy detailed in Sections \ref{sec:conventional_memory_replay}, \ref{sec:adrb} and \ref{sec:train_obj}.

\subsection{Motivation}
\label{sec:motivation}
ADRM prevents memory overfitting in the CL model by diversifying and enhancing the complexity of memory samples using an adversarial FGSM approach \cite{wong2019fast}. It introduces varying strengths of adversarial perturbations to each class memory sample to enrich diversification within the class (see Figure \ref{fig:main}) using one-step FGSM adversarial. The resultant diversified memory samples are more complex and challenging to overfit by the CL model. Each step of adversarial diversification results in a unique diversified sample \cite{wong2019fast}. Additionally, the adversarial diversification of the rehearsal memory reinforces the CL model to minimize feature distribution drift within each memory class, better learn intra-class boundaries, resulting in generalization and robustness, and reduce catastrophic forgetting.

\subsection{Conventional Memory Replay}
\label{sec:conventional_memory_replay}
The standard rehearsal approach, also known as experience replay \cite{chaudhry2019continual}, involves combining a limited number of memory samples from previous tasks with mini-batches of the current task dataset. This approach minimizes data risk for the rehearsal memory ($\mathcal{M}$) and the current task data ($\mathcal{D}_t$). Formally, the optimization process can be expressed as follows:

\begin{equation}
\label{eq:conventional_rehearsal}
\min_{\theta \in \Theta} \left[ \mathbb{E}_{(x_t,y_t)\sim\mathcal{D}_t} \mathcal{L}(\theta, x_t, y_t) + \mathbb{E}_{(x_m,y_m)\sim\mathcal{M}_t} \mathcal{L}(\theta, x_m, y_m) \right], 
\end{equation}
where $t$ represents the $t^{th}$ task or step, $\theta$ denotes the parameters of the CL model, and $\mathcal{L}$ represents the loss function, typically cross-entropy loss, commonly used, $\mathcal{D}_t$ represents the current task data, while $\mathcal{M}_t$ represents the rehearsal memory data for previous tasks. The pairs $(x_t, y_t)$ and $(x_m, y_m)$ are samples, drawn from $\mathcal{D}_t$ and $\mathcal{M}_t$, respectively. Given the memory constraints, rehearsal memory can only store a limited subset of old data. This limitation increases the model's tendency to overfit these memory samples \cite{de2021continual, jin2021gradient}. As a result, the effectiveness of these samples in mitigating forgetting is diminished \cite{de2021continual, jin2021gradient}. Furthermore, this memory overfitting impairs the model's generalization capabilities and intensifies the catastrophic forgetting of previously learned tasks.

Our proposed ADRM addresses the rehearsal memory overfitting by enhancing the diversity and complexity of the rehearsal memory samples in an adversarial way, thereby preventing the model from overfitting the memory samples. The details of this approach are explained in further detail in the next sections (\ref{sec:fgsm_diversification}, and \ref{sec:adrb} ).

\subsection{Adversarial Diversification of Rehearsal Memory Using FGSM}
\label{sec:fgsm_diversification}
Formally, the ADRM employs the one-step FGSM on the rehearsal memory ($\mathcal{M}$) for performing memory sample diversification \cite{wong2019fast, shafahi2019adversarial}, is represented as follows.

\begin{equation}
\label{eq:fsgm}
x_{\text{diversified}} = x_m + \epsilon \cdot \text{sign}(\nabla_{x_{m}} J(\theta, x_m, y_m)), 
\end{equation}
where $x_{\text{diversified}}$ represents the adversarially diversified samples of the corresponding $x_m$ memory sample, and epsilon ($\epsilon$) is the perturbation strength, the $\text{sign}$ function is used to determine the direction of the perturbation, while $\nabla_{x_{m}} J(\theta, x_m, y_m)$ signifies the gradient of the loss function with respect to the input $x_m$. Here, $\theta$ denotes the parameters of the CL model, and $y_m$ is the true label for $x_m$. The pairs $(x_m, y_m)$ belongs to the rehearsal memory ($\mathcal{M}$). The $x_{\text{diversified}}$ represents the adversarially diversified samples (with increased complexity) of the $x_m$ original memory samples. These samples are interleaved with mini-batches sampled from the current task data ($\mathcal{D}_t$). Notably, the adversarially diversified memory samples created at each CL step are unique and can be generated in any desired number for each previously learned class (i.e., memory samples) through a single FGSM step \cite{wong2019fast, shafahi2019adversarial}

\subsection{Creating Diversified Rehearsal Memory}
\label{sec:adrb}
We randomly select a mini-batch of data from the rehearsal memory ($\mathcal{M}$) and employ the FGSM to generate adversarially perturbed/diversified memory samples. The diversified samples are divided into subsets $\mathcal{M}_{d}$ and $\mathcal{M}_{b}$; the former subset
contains those diversified memory samples where the adversarial perturbation led to misclassification, while the latter subset
contains diversified memory samples that were still classified
correctly by the CL model despite the perturbation. Together, these subsets of diversified memory samples (i.e., $\mathcal{M}_{d}$) and $\mathcal{M}_{b}$)  can be referred to as a diversified version of rehearsal memory (i.e., $\mathcal{M}$). The $\mathcal{M}_{d}$ subset increases diversity and reduces overfitting, while the $\mathcal{M}_{b}$ subset helps maintain decision boundaries between classes and ensures
robust continual learning with stable feature distributions, resulting in less catastrophic forgetting (see Figure \ref{fig:main}).


\subsection{Memory Diversification Ratios in ADRM}
\label{sec:train_obj}
In order to study the impact of memory diversification on the CL model, we incorporated adversarially diversified rehearsal memory in five different ratios: $10\%$, $25\%$, $50\%$, $75\%$, and $100\%$. For example, an ADRM model that rehearses $10\%$ of both $\mathcal{M}_{d}$ and $\mathcal{M}_{b}$ along with the original memory samples, would be referred to as ADRM ($0.1$). Similarly, ADRM models trained with $25\%$, $50\%$, $75\%$, and $100\%$ are denoted as ADRM ($0.25$), ADRM ($0.5$), ADRM ($0.75$), and ADRM ($1$), respectively.


\section{Experimental setup}
\subsection{Datasets} 
We utilized the widely-used split CIFAR10 benchmark dataset \cite{krizhevsky2009learning, zhao2020maintaining, wu2019large, de2021continual, zhou2023revisiting}. CIFAR10 is known for its complexity and consists of ten distinct classes, each containing $6,000$ images. Among these images, $5,000$ were used for training, while the remaining $1,000$ were reserved for the test set \cite{krizhevsky2009learning}. We also evaluated the robustness of CL models against natural noise using the CIFAR10-C dataset  \cite{hendrycks2018benchmarking}.  Additionally, we employed the adversarially perturbed CIFAR10 dataset to evaluate the robustness of CL models against adversarial noise \cite{khan2022adversarially}. Figure \ref{fig:acc_adv_attack} presents sample images from the adversarially perturbed CIFAR10 dataset \cite{khan2022adversarially}.

\subsection{Protocols}
We utilized the challenging class-incremental learning (CIL) protocol, which simulates real-life situations where the model learns sequentially from streaming tasks without prior task identification \cite{de2021continual, masana2022class, wang2023comprehensive, van2019three}. According to this protocol, the CIFAR-10 dataset is divided into nine, five, and two tasks, each with a fixed memory size of 1024 examples, following the CIL protocol \cite{lopez2017gradient, masana2022class, van2019three}.

\begin{table}[ht]
\caption{Comparative performance of CL methods on the Split-CIFAR10 dataset over 9 steps, 5 steps, and 2 steps. The proposed ADRM method, rehearsing with $10\%$ adversarially diversified memory, demonstrates superior performance over variants incorporating $25\%$, $50\%$, $75\%$, and $100\%$ diversified memory. The ADRM outperforms several established CL methods and achieves comparable results to state-of-the-art CL approaches. Each experiment was repeated five times with different random seeds to ensure the robustness of our findings.}
\label{table:accuracy}
\centering
\begin{tabular}{lccccc}
\hline  
\noalign{\vskip 0.05cm}
\textbf{CL Methods} &  \multicolumn{3}{c}{\textbf{Split-CIFAR10}} \\
&  \textbf{9 steps} &  \textbf{5 steps} &  \textbf{2 steps}  \\ \hline 
\noalign{\vskip 0.025cm}
Joint & \multicolumn{3}{c}{94.8} \\
Fine-tune & 11.11 & 18.54 & 45.31\\ 
\hline
\noalign{\vskip 0.025cm}
Experience Replay \cite{rolnick2019experience} & 67.62   & 75.97   & 80.26   \\  
iCaRL \cite{rebuffi2017icarl} & 69.25  & 74.85   & 79.98   \\ 
BiC \cite{wu2019large} & 53.11   & 71.01  & 86.57  \\  
PODNet \cite{douillard2020podnet} & 72.41   & 76.96   & 87.64   \\  
WA \cite{zhao2020maintaining} & 75.56   & 81.67   & 85.34   \\  
DER \cite{yan2021dynamically} & 74.33   & 78.14   & 82.57   \\  
SimpleCIL \cite{zhou2023revisiting} & 52.67   & 54.54   & 75.01   \\  
FOSTER \cite{wang2022foster} & 74.61   & 80.40   & 83.89   \\  
FETRIL \cite{petit2023fetril} & 65.26   & 67.51   & 84.52   \\  
MEMO \cite{zhou2022model} & 80.60   & 85.93   & 87.81   \\  
ADRM (0.1)  & 74.76   & 80.59   & 83.95   \\  
ADRM (0.25) & 72.39   & 79.15
   & 85.39   \\  
ADRM (0.5) & 69.20   & 76.14   & 81.41   \\  
ADRM (0.75)  & 68.59   & 76.36
   & 80.17   \\  
ADRM (1)  & 65.39   & 76.21   & 82.86   \\  
\hline
\end{tabular}
\end{table}
 
\subsection{Baselines}
We conducted a comprehensive comparative analysis of our proposed method with the existing well-established and state-of-the-art CL methods, which included:  1) Experience Replay (ER), also referred to as Standard Rehearsal \cite{chaudhry2019tiny}. ER employs reservoir sampling to store a compact subset of data from previous tasks. During learning of the new tasks, a random subset of examples from the memory buffer is sampled and combined with the incoming mini-batch data to mitigate forgetting; 2). Incremental Classifier and Representation Learning (iCaRL) incrementally train a model on new classes while retaining knowledge of previous ones through representation learning and an exemplar memory mechanism \cite{rebuffi2017icarl}; 3). Bias Correction (BiC) that adjusts a model's bias to balance performance between newly added and original classes \cite{wu2019large}; 4). Pooling-based Online Distillation Network (PODNet) addresses catastrophic forgetting by utilizing spatial-based distillation and feature pooling \cite{douillard2020podnet}; 5). Weight Alignment (WA) uses knowledge distillation to maintain the discrimination within old classes adjust, followed by correcting the biased weights in the fully connected layer of the model  \cite{zhao2020maintaining}; 6). Dynamically Expandable Representation (DER) learns and dynamically expands the features representation with a new extractor, which is followed by re-training the classifier using currently available data to reduce bias in the classifier weight caused by imbalanced training \cite{yan2021dynamically}; 7). SimpleCIL uses pre-trained model embeddings and sets classifier weights directly to prototype features of each class, avoiding the need for additional downstream task training\cite{zhou2023revisiting}; 8). Feature BoOSTing and CompreEssion for class-incRemental learning (FOSTER) introduces a two-stage method that expands and then compresses a model for class-incremental learning \cite{wang2022foster}, which effectively balances new-category learning with preserving old knowledge; 9). FeTrIL, Similar to FOSTER, FeTrIL introduces a two-stage method for class-incremental learning, encompassing both model expansion and compression stages \cite{petit2023fetril}, which effectively balances new category learning with the preservation of prior knowledge; 10). Memory-efficient Expandable MOdel (MEMO) optimizes memory efficiency by expanding specialized layers for new tasks while sharing generalized layers, ensuring an efficient and effective approach to continual learning \cite{zhou2022model}. We also included a fine-tuned model that learned all tasks in sequence without specific strategies to prevent forgetting. We also used a joint model trained on the entire dataset, which included all available classes. The fine-tuned model established the lower performance, whereas the joint model established the upper bound performance.

\begin{figure*}
\centering
\includegraphics[width=0.85\textwidth]{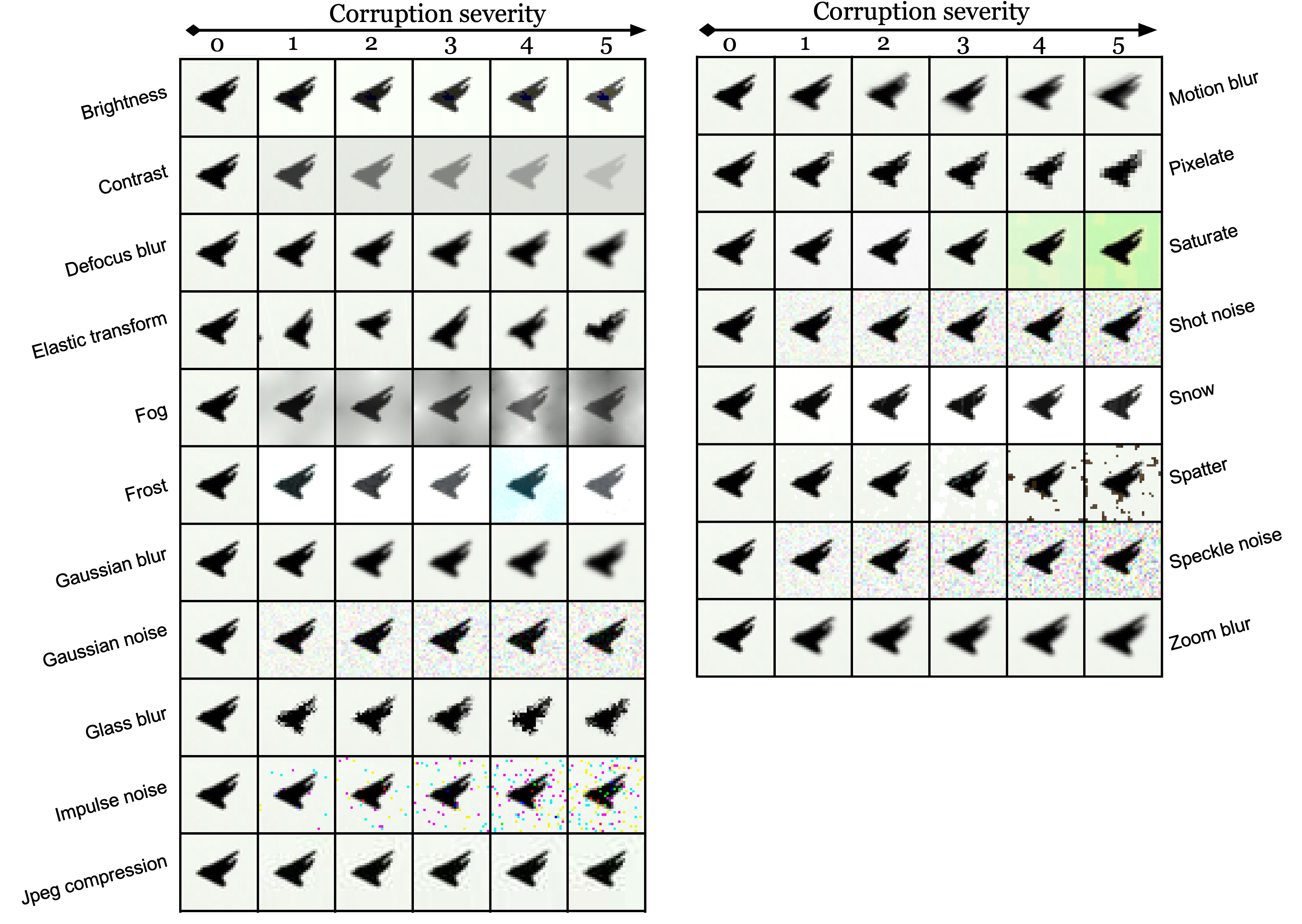}
\caption{Examples of the six different severity levels for the 19 different noise types. Severity 0 represents the original clean image whereas severity 5 corresponds to the most severe corruption (Best viewed in color).}
\label{fig:corrupted_images}
\end{figure*}

\begin{figure*}
\centering
\includegraphics[width=1\textwidth]{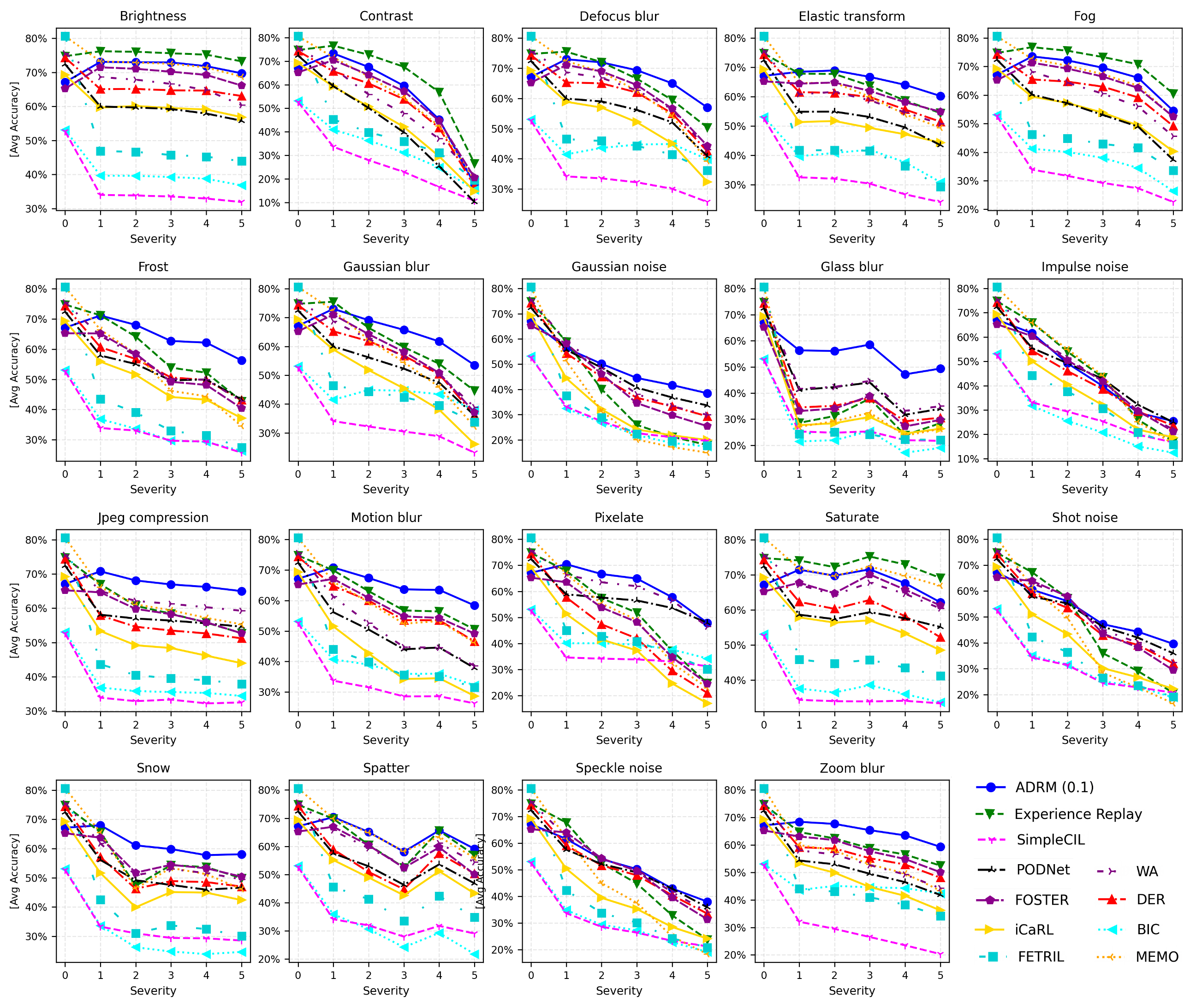} 
\caption{Robustness comparison of eleven CL approaches against nineteen types of natural image corruptions. Each subplot corresponds to a different noise type, with the x-axis representing the severity level from 0 (no corruption) to 5 (maximum corruption) and the y-axis displaying the average accuracy of each model. The ADRM shows resilience and comparatively suffers less from catastrophic forgetting. It outperformed other models in 15 out of the 19 tested noise conditions, highlighted with a yellow rectangle in the graph (best viewed in color).}
\label{fig:acc_noise}
\end{figure*}

\subsection{Training details}
We used the PyTorch framework \cite{paszke2017automatic} and PyCIL codebase \cite{zhou2023pycil} to implement our experiments. We employed the standard ResNet32 architecture with Xavier initialization \cite{he2016deep, glorot2010understanding} and a batch size of $256$. We used a stochastic gradient descent optimizer to train the models with an initial learning rate of $0.01$, a momentum of $0.9$, and a learning rate decay of $0.1$. We used the standard cross-entropy loss function for classification loss. The training epochs were set to $200$ for the initial task and $128$ for subsequent tasks. We also used standard data augmentation techniques, such as random flip, cropping, and adjustments in brightness and contrast \cite{alomar2023data}. We set the hyper-parameters specific to each baseline model to their default configurations as specified in their respective papers \cite{rebuffi2017icarl, douillard2020podnet, yan2021dynamically, zhao2020maintaining, wang2022foster, wu2019large} and implemented in PyCIL \cite{zhou2023pycil} to ensure optimal performance for each baseline for fair comparison. To diversify memory samples, we utilized a perturbation parameter (epsilon $\epsilon$) in the FGSM attack. We randomly chose epsilon ($\epsilon$) from a uniform distribution between $\frac{1}{255}$ and $\frac{16}{255}$. This range was carefully selected to create adversarial perturbations intense enough to diversify memory samples while preserving the visual characteristics of the original memory class.

\subsection{Evaluation}
We evaluate the performance of the CL models using the widely adopted metric of average classification accuracy (ACA) \cite{wang2023comprehensive, douillard2022dytox, rebuffi2017icarl, petit2023fetril, wang2022foster, lopez2017gradient}, which is calculated by assessing the final trained model on all the learned tasks. Formally, ACA is defined as:
\begin{equation}
    ACA = \frac{1}{T} \sum_{i=1}^{T}R_{T,i},
\end{equation}
where $R$ represents accuracy, $T$ is the total number of tasks, and $i$ is the task index.

\begin{figure*}
\centering
\includegraphics[width=0.99\textwidth]{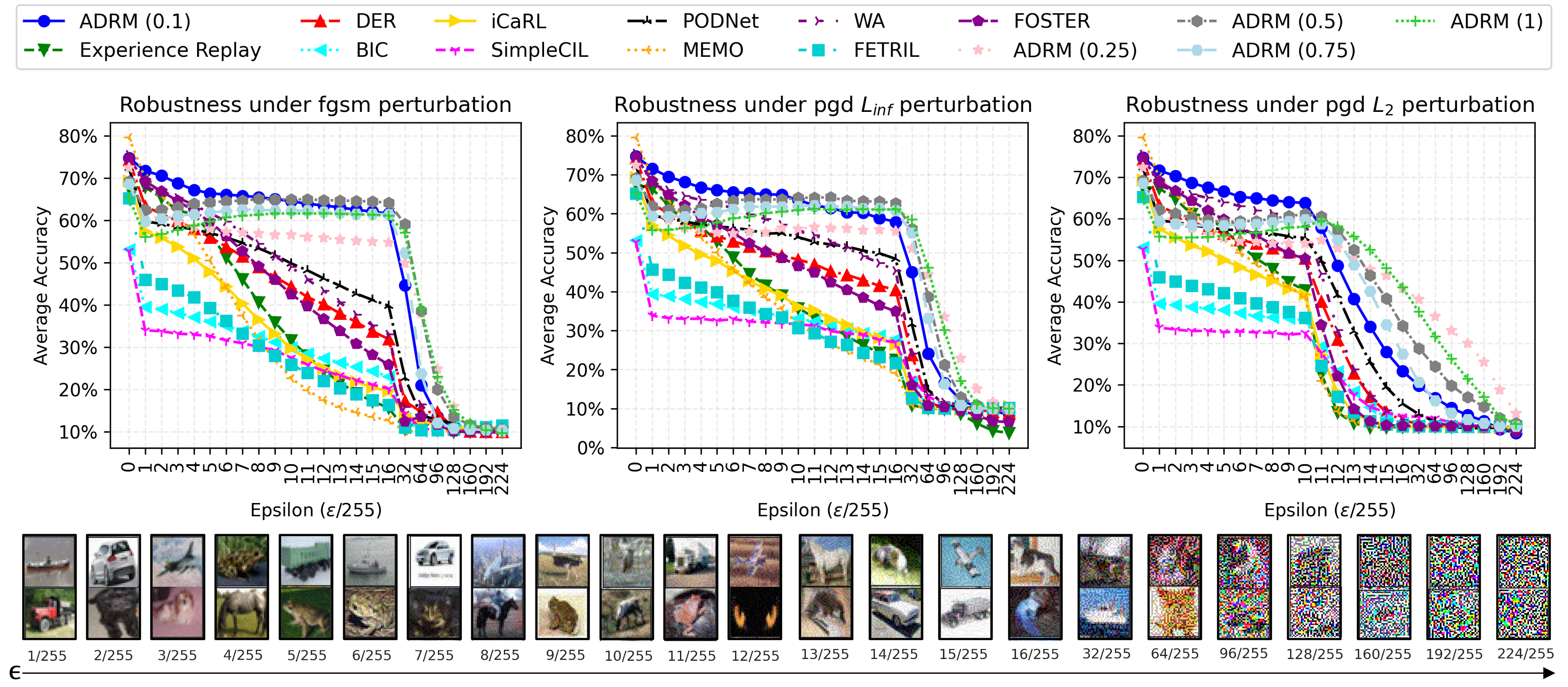}
\caption{Robustness comparison of fifteen CL approaches on the adversarially perturbed CIFAR-10 validation set \cite{khan2022adversarially}. The x-axis represents the epsilon value ($\epsilon/255$), indicating the strength of the adversarial perturbation, while the y-axis shows the models' average accuracy. The sample images at the bottom are adversarially perturbed using the Fast Gradient Sign Method (FGSM) with increasing perturbation strength ($\epsilon$) from left to right. ADRM and its variants demonstrate enhanced adversarial robustness over a range of attack intensities relative to other CL approaches, indicating a stronger grasp of the fundamental class concepts within the dataset (best viewed in color).}
\label{fig:acc_adv_attack}
\end{figure*}

\section{Results and Discussion}

\begin{figure*}
\centering
\includegraphics[width=0.80\textwidth]{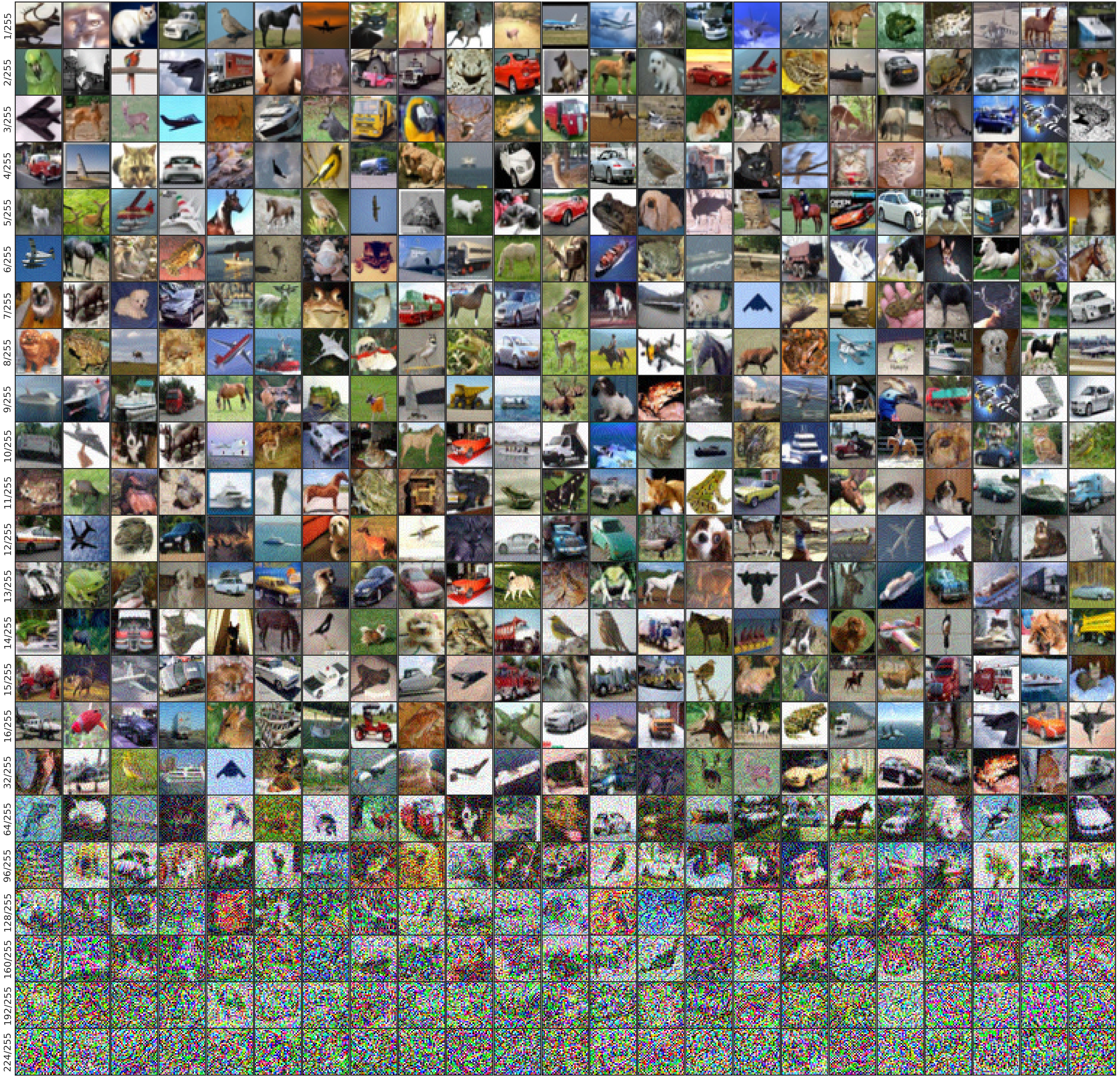}
\caption{Adversarial examples created using PGD-$L_{\inf}$. On the $x$-axis, we can see the different adversarially perturbed images from the dataset. The different strengths of the level of the attacks ($\epsilon$) are plotted in the $y$-axis. The level of the attack ($\epsilon$) determines the strength of the adversarial attack; the higher the value, the stronger the adversarial attack and the more perturbed the input image (Best viewed in color).}
\label{fig:adv_images} 
\end{figure*}

\begin{figure*}
\centering
\includegraphics[width=0.80\textwidth]{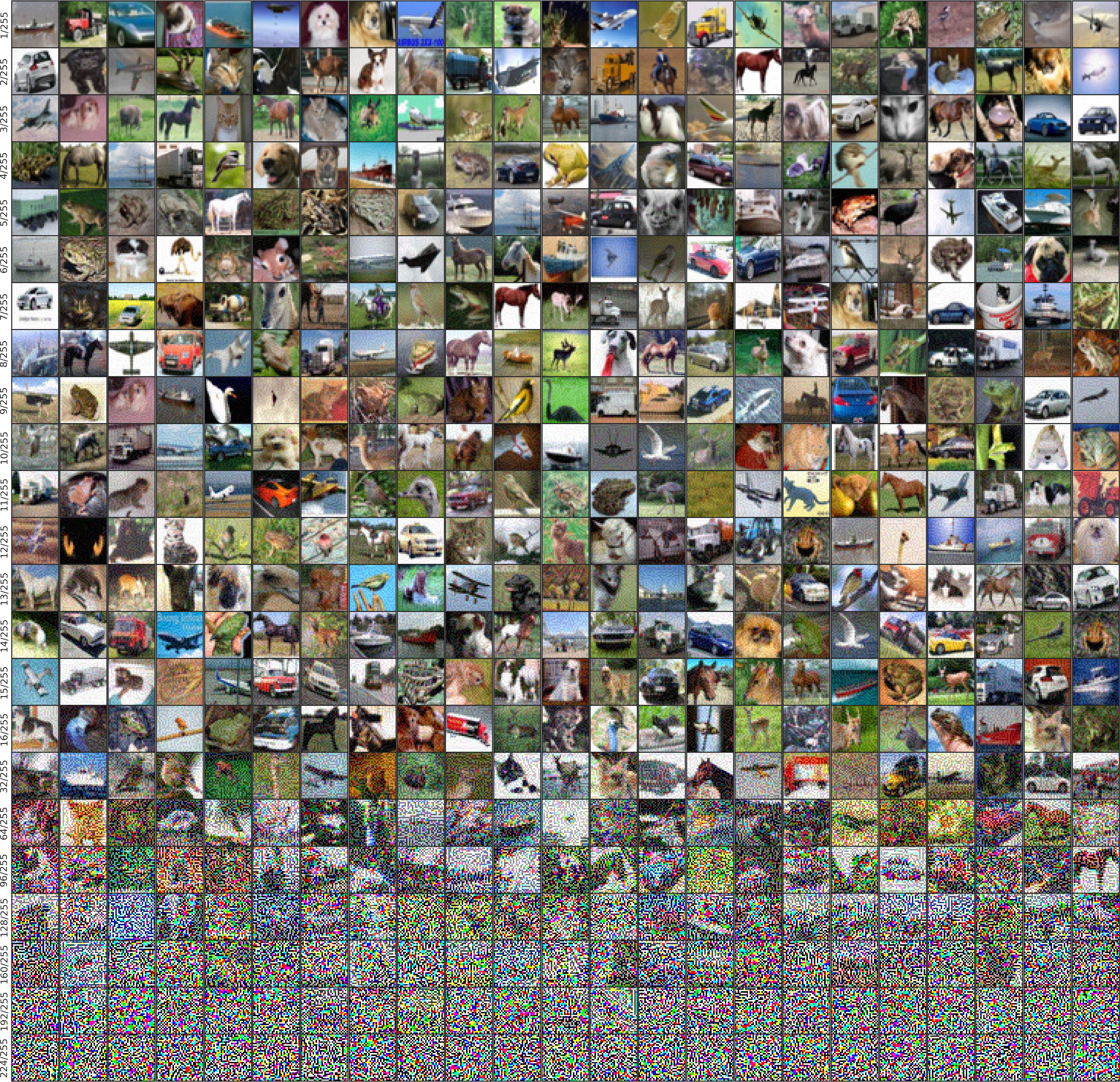}
\caption{Adversarial examples created using FSGM attack. On the $x$-axis, we can see the different adversarially perturbed images from the dataset. The different strengths of the level of the attacks ($\epsilon$) are plotted in the $y$-axis. The level of the attack ($\epsilon$) determines the strength of the adversarial attack; the higher the value, the stronger the adversarial attack and the more perturbed the input image (Best viewed in color).}
\label{fig:pgd_adv_images}
\end{figure*}




Table \ref{table:accuracy} presents a comparative performance analysis of the ADRM and its variants methods with baseline CL approaches for tasks involving nine, five, and two task configurations on the CIFAR10 dataset. Notably, the ADRM variant with a $10\%$ inclusion ratio of diversified memory samples, referred to as ADRM ($0.1$), performed the best with the highest average accuracy among all the other variants of ADRM. Moreover, the ADRM ($0.1$) variant demonstrated superior performance compared to several well-established methods, such as Experience Replay, iCaRL, BiC, PODNet, SimpleCIL, and FetRIL, especially in situations that involve nine tasks. Additionally, ADRM obtained comparable average accuracy to methods like WA, DER, and FOSTER. 

\subsection{Balancing Generalization and Robustness: The Impact of Adversarially Diversified Memory Samples in ADRM Model Training}
Recall that the different variants of ADRM rehearse various proportions/ratios of adversarially diversified memory samples, such as $10\%$, $25\%$, $50\%$, $75\%$, and $100\%$  of rehearsal memories (which were represented by $\mathcal{M}_d$ and $\mathcal{M}_b$ in Section \ref{sec:train_obj}). We conducted extensive experiments and found a tradeoff between the generalization and robustness of ADRM models. Specifically, as we increased the proportion of diversified memory samples, the model's robustness improved, but its ability to generalize to new task data was adversely affected. Surprisingly, we observed that the optimal balance was achieved using a $10\%$ inclusion ratio of diversified memory samples. However, when gradually increasing the ratio of diversified memory samples from $10\%$ to $100\%$, we noticed a decline in performance, especially in the 9-task scenario (see Table \ref{table:accuracy}). Furthermore, our findings suggest that rehearsing with a $10\%$ inclusion ratio of diversified memory samples enhances generalization and robustness and prevents memory overfitting, resulting in less catastrophic forgetting in the CL model (discussed in the following Section \ref{section:robustness_in_cl}). 


\subsection{Diversifying Rehearsal Memory Leads to Robust Continual Learning Models}
\label{section:robustness_in_cl}
The ADRB achieved lower average accuracy in comparison to the highest-performing MEMO approach (refer to Table \ref{table:accuracy}). However, it demonstrated stability and experienced less catastrophic forgetting under natural and adversarial conditions as compared to the CL baselines.

\subsubsection{Evaluating robustness against common corruptions}
Figure \ref{fig:acc_noise} shows the performance of different CL methods against nineteen different types of noise (see Figure \ref{fig:corrupted_images} for samples images from CIFAR10-C dataset). The ADRM model was the most resilient and least prone to forgetting, maintaining the highest average accuracy in $15$ out of $19$ noise types. However, it is worth noting that the MEMO model performed the best overall. Nevertheless, it struggled with natural noise and showed a higher tendency to catastrophic forgetting than the ADRM model and other established CL methods.

\subsubsection{Evaluating robustness against adversarial corruptions}
Figure \ref{fig:acc_adv_attack} illustrates that ADRM variants consistently outperformed the CL baseline methods in performance across three different versions of the dataset, each altered using different adversarial attack methods: PGD-$L_{\inf}$, PGD-$L_{2}$, and FGSM (see Figures \ref{fig:pgd_adv_images} and \ref{fig:adv_images} for samples images from adversarially perturbed CIFAR10 datasets). Notably, the highest-performing CL approach, MEMO, exhibited worse performance and the highest forgetting compared to other CL approaches. Interestingly, the ADRM variant with a $10\%$ diversity ratio attained the best results among its peers while maintaining accuracy similar to well-established CL approaches. These observations highlight the effectiveness of adding even a small amount of adversarial diversification (i.e., $10\%$) in bolstering the resilience of CL models and resulting in less forgetting under adversarial conditions.

\begin{figure*}
\centering
\includegraphics[width=1\textwidth]{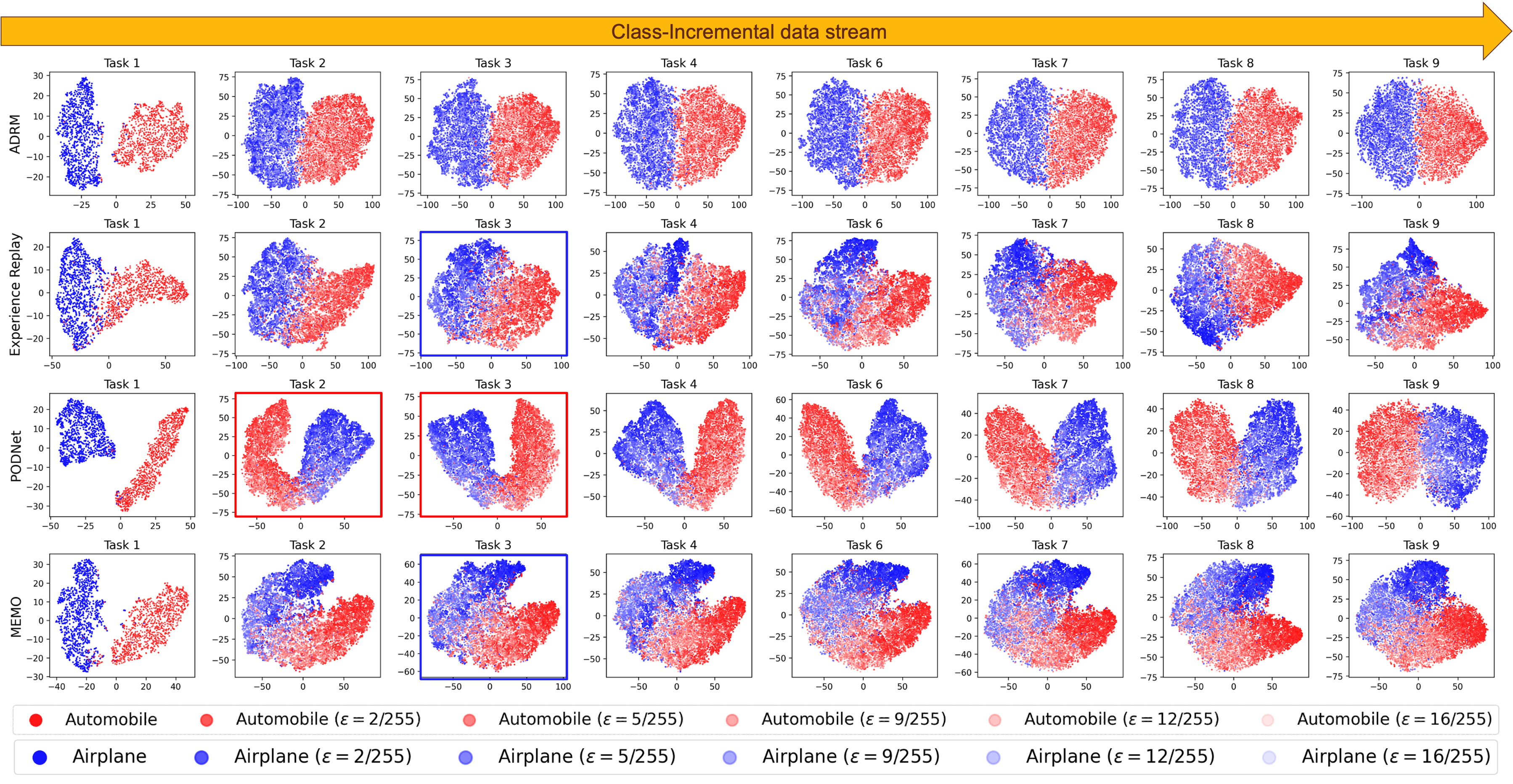}
\caption{
Presents t-SNE visualizations of feature distributions for 'airplane' and 'automobile' classes in the CIFAR-10 dataset, under adversarial perturbations with $\epsilon$ values $\frac{2}{255}$, $\frac{5}{255}$, $\frac{3}{255}$, $\frac{9}{255}$, $\frac{12}{255}$ and $\frac{16}{255}$ across nine learning tasks. The original 'airplane' and 'automobile' class samples are denoted by dark blue and red dots, respectively, with lighter shades representing increased levels of adversarial perturbation. The sub-figures a), b), c), and d) display t-SNE visualizations for ADRB, ER, PODNet, and MEMO models, respectively, using features extracted from the last layer of ResNet32. The first column in each sub-figure presents the feature distribution for the initial two-class (airplane and automobile) learning task, highlighting the distinct separation achieved by all models. The subsequent columns illustrate each model's feature distribution evolution after learning subsequent tasks. In sub-figure a), Throughout the nine task learning, the ADRM model shows stable feature distributions for the 'airplane' and 'automobile' classes, with no shifts and minimum overlap. The other models exhibit an increasing trend of high overlap between the feature distributions (highlighted in a blue rectangle) and also suffer from feature distribution shifts (highlighted in a red rectangle). PODNet in sub-figure c) is the second-best performer, but it suffers from feature distribution shifts. On the other hand, MEMO in sub-figure d) suffers from the highest overlapping of feature distributions, which indicates the highest forgetting and lower performance in adversarial conditions (best viewed in color).
}
\label{fig:t_sim_viz}
\end{figure*}

\subsection{Diversifying Rehearsal Memory Leads to Stable Feature Distribution}
We analyzed the feature distributions learned by various CL approaches to gain insights into the catastrophic forgetting of previously learned classes using the adversarially perturbed CIFAR10 dataset (see Figure \ref{fig:t_sim_viz}) \cite{khan2022adversarially}. We compared the feature distributions of four CL approaches, ADRM, Experience Replay, PODNet, and MEMO, across nine tasks for `airplane' and `automobile' classes using pre-softmax logits of the ResNet32 architecture. Our findings indicate that ADRM models learned stable feature distributions with minimal drifts and class overlap, while the other approaches had noticeable drifts and increasing overlap. Overall, Figure \ref{fig:t_sim_viz} shows that ADRM is more effective in learning core concepts and intra-class boundaries, reducing class forgetting even in challenging adversarial conditions.

\subsection{Analyzing Feature Similarity in Continual Learning Models Using Central Kernel Alignment (CKA)}
\begin{figure*}[htbp]
\centering
\includegraphics[width=0.85\textwidth]{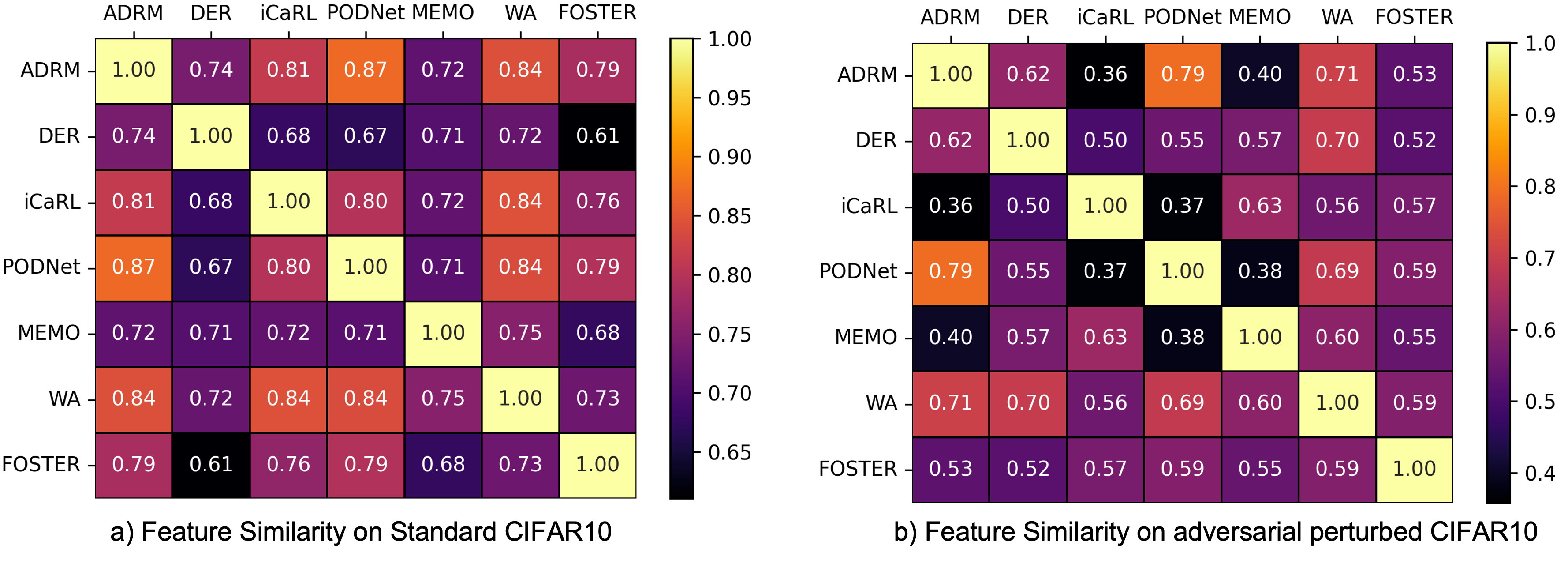}
\caption{Features similarity matrices calculated using centered kernel alignment (CKA). 
On the standard CIFAR-10, the ADRM and the other CL approaches exhibit high feature similarity in learned representation. In contrast, low feature similarities are observed on the adversarially perturbed CIFAR-10 dataset. PODNet, DER, and WA, with high similarity scores to ADRM, also exhibit enhanced performance in adversarial conditions and suffer less catastrophic forgetting. In contrast, models with lower feature similarity to ADRM suffer more in adversarial conditions, highlighting the crucial trade-off between robustness, generalization, and catastrophic forgetting within the CL domain. The matrices point towards the continually adversarially robust features that enhance the overall robustness of CL models and prevent catastrophic forgetting(best viewed in color).}
\label{fig:sim_matrix}
\end{figure*}

Figure \ref{fig:sim_matrix} illustrates the similarity matrices of the learned feature representations for seven CL approaches, computed on the validation sets of the CIFAR10 and adversarially perturbed CIFAR10 datasets \cite{khan2022adversarially}, computed using CKA \cite{kornblith2019similarity}. We observed high similarities in the learned feature representations among ADRM and various CL approaches computed on the standard CIFAR10 dataset. In contrast, We observed lower similarities in the learned feature representations among the ADRM and various CL approaches computed on the adversarially perturbed CIFAR10 dataset. In addition, models with higher feature similarity scores to ADRM, such as DER and PODNet, demonstrated better resistance to adversarial conditions and reduced catastrophic forgetting. Conversely, models with low feature similarity scores to ADRM's features, such as iCaRL and MEMO, showed increased vulnerability to adversarial conditions and greater forgetting of previously learned classes, indicating limitations in their learned representations when faced with adversarial conditions.

\subsection{Learned Feature Disentanglement}

We used a feature disentanglement framework from \cite{ilyas2019adversarial} to understand the feature representations learned by CL models. ADRM learned salient features of the automobile, such as tires and the frame, highlighting its focus on the class's key attributes. Other models, such as Experience Replay, PODNet, and MEMO, exhibited a more scattered and less class-specific feature pattern, making them vulnerable to noise and causing a higher forgetting of previously learned classes.

\begin{figure}
\centering
\includegraphics[width=.45\textwidth]{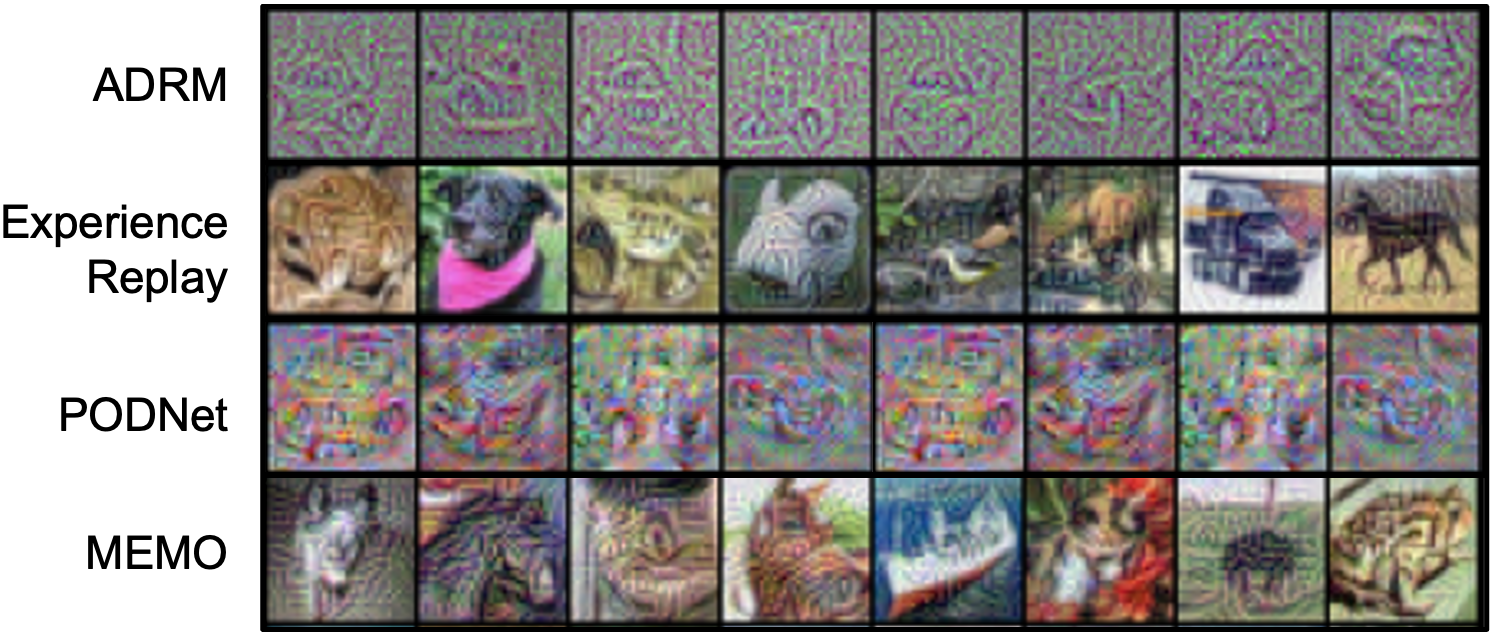}
\caption{Visualizations of disentangled features for the 'automobile' (or 'car') class using the framework presented in \cite{ilyas2019adversarial} in the input space. \nth{1} row presents the ADRM learned features that concentrate on salient class features, such as tires and body frames, resulting in less catastrophic forgetting in adversarial conditions. On the other hand, \nth{2}, \nth{3}, and \nth{4} rows present the features for Experience Replay, MEMO models, comprising incoherent activations and non-class-specific artifacts, aligning with their lower performances in adversarial scenarios (best viewed in color).}
\label{fig:features_viz}
\end{figure}

\section{conclusion}

In this paper, we introduce the Adversarially Diversified Rehearsal Memory (ADRM) approach to address the challenges of rehearsal memory overfitting and catastrophic forgetting challenges in continual learning. ADRM employs the Fast Gradient Sign Method (FGSM) to introduce memory diversification, aiming to accomplish two primary goals. Firstly, it enhances the rehearsal memory's diversity to mitigate memory overfitting. Our extensive experiments on the CIFAR10 dataset demonstrate that ADRM surpasses several CL approaches and achieves comparable performance to state-of-the-art CL approaches. Secondly, ADRM enhances the robustness of the CL model, leading to reduced forgetting of previously learned classes under both natural and adversarial conditions. Specifically, ADRM maintained the highest average accuracy in 15 out of 19 noise types (e.g., Gaussian noise and impulse noise), as evaluated on the CIFAR10-C dataset. Moreover, ADRM demonstrated superior resilience in adversarial conditions compared to other CL models, with less forgetting of previously learned classes. We also performed t-SNE visualizations and employed centered kernel alignment (CKA) similarity scoring to investigate the stability of feature distributions, as well as to understand the learned feature representations and their similarities among various CL approaches. We observed that models with features more aligned with ADRM (i.e., exhibiting a higher similarity score) demonstrated enhanced robustness to both natural and adversarial noises and experienced reduced catastrophic forgetting. These insights deepen our understanding of feature representation learning within the realm of continual learning, underscoring the importance of continually-robust feature learning for CL models to avert catastrophic forgetting.

\section{Acknowledgement}
This work was supported by National Science Foundation (NSF) Award NSF OAC 2008690

\bibliographystyle{IEEEtran}

\bibliography{bibliography}{}

\end{document}